\documentclass[twocolumn,english,american]{svjour3}
\usepackage[T1]{fontenc}
\usepackage[utf8]{inputenc}
\usepackage{url}
\usepackage{amsmath}
\usepackage{amssymb}
\usepackage{graphicx}

\makeatletter

\providecommand{\tabularnewline}{\\}

\newenvironment{svmultproof}{\begin{proof}}{\qed\end{proof}}

\RequirePackage{fix-cm}

\smartqed  

\usepackage{physics}
\usepackage{babel}

\makeatother

\usepackage{babel}
\addto\captionsamerican{}
\addto\captionsamerican{}
\addto\captionsamerican{}
\addto\captionsamerican{}
\addto\captionsamerican{}
\addto\captionsamerican{}
\addto\captionsenglish{}
\addto\captionsenglish{}
\addto\captionsenglish{}
\addto\captionsenglish{}
\addto\captionsenglish{}
\addto\captionsenglish{}

\begin{document}

\title{Geometrical analysis of polynomial lens distortion models\thanks{This work has been partially supported by the Ministerio de Economí­a,
Industria y Competitividad (AEI/FEDER) of the Spanish Government under
projects TEC2016-75981 (IVME) and TIN2016-75982-C2-2-R (HEIMDAL-UPM). This is a pre-print of an article published in the Journal of Mathematical Imaging and Vision. The final authenticated version is available online at: https://doi.org/10.1007/s10851-018-0833-x}}

\titlerunning{Polynomial lens distortion models}

\author{José I. Ronda, Antonio Valdés}

\authorrunning{Short form of author list}

\institute{José I. Ronda \at Grupo de Tratamiento de Imágenes \\
 Universidad Politécnica de Madrid\\
 \email{jir@gti.ssr.upm.es}\\
Antonio Vald\'es \at Departamento de \'Algebra, Geometr\'{\i}a
y Topolog\'{\i}a\\
Universidad Complutense de Madrid\\
\email{avaldes@ucm.es}}

\date{Received: date / Accepted: date}
\maketitle
\begin{abstract}
Polynomial functions are a usual choice to model the nonlinearity
of lenses. Typically, these models are obtained through physical analysis
of the lens system or on purely empirical grounds. The aim of this
work is to facilitate an alternative approach to the selection or
design of these models based on establishing a priori the desired
geometrical properties of the distortion functions. With this purpose
we obtain all the possible isotropic linear models and also those
that are formed by functions with symmetry with respect to some axis.
In this way, the classical models (decentering, thin prism distortion)
are found to be particular instances of the family of models found
by geometric considerations. These results allow to find generalizations
of the most usually employed models while preserving the desired geometrical
properties. Our results also provide a better understanding of the
geometric properties of the models employed in the most usual computer
vision software libraries.

\subclass{51 \and 78} 
\end{abstract}

\section{Introduction\label{sec:Introduction}}

The correction of lens distortion is a relevant problem in computer
vision and photogrammetry \cite{Hartley-Zisserman}. Lens distortion
models the departure of the image capturing device from the theoretical
pin-hole model and consists essentially in an image warping process. 

Most of the proposed lens distortion models are given by an analytical
expression of the space variables and the model parameters, although
some efforts have also being made in order to depart from concrete
analytical expressions \cite{Hartley}. These closed-form
expressions usually express the position of the distorted points as
a function of the ideal undistorted points given by the pinhole assumption,
although in some cases it is the inverse of this function what is
given by the model functions \cite{Claus}. 

Lens distortion models can either result from the analysis of the
physical problem or from a pragmatic approach led by the empirical
capacity of the model to fit the observed data and the existence of
practical algorithms to compute the model parameters. The concrete
parameters of the distortion function are frequently computed within
the bundle-adjustment process of a 3D scene reconstruction \cite{Claus,Weng,Li},
but it is often possible to obtain these parameters from a single
image that contains an element of known geometry, such as a calibration
grid or a set of lines \cite{Alvarez,Strand,Wu,Devernay}.

The first and probably most employed analytical form of lens distortion
models is given by polynomials \cite{Conrady,Brown,Weng}. A natural
generalization is that of rational functions \cite{Claus}, although
some empirical studies \cite{Tang} attribute a similar modeling capabilities
to both approaches.

A large part of the literature on these models assumes a radial rotationally
invariant (RRI) distortion function \cite[p. 191]{Hartley-Zisserman}.
This strong geometrical requirement stems from the assumption that
the capturing system is a rotationally symmetric structure. While
these models suffice for some applications, those requiring higher
precision must also account for such phenomenons as the non-alignment
of the axes of the lens surfaces or the lack of paralellism of the
lens and the imaging surface. The first is usually addressed by the
decentering lens distortion model \cite{Conrady} and the second by
means of the thin prism model \cite{Brown}. The model employed in
the computer vision software library OpenCV \cite{OpenCV} integrates
a rational term to model radial rotationally invariant distortion
with polynomial terms accounting for thin prism and decentering distortion.

Radial rotationally invariant distortion, decentering distortion and
thin-prism distortion are examples of models with interesting geometrical
properties. They are linear, in the sense that the models constitute
a vector space, they are isotropic, i.e., invariant to plane coordinate
rotation and, from physical considerations, are formed of functions
that are reflection-symmetric with respect to some axis. Some questions
arise naturally:
\begin{itemize}
\item Are decentering and thin-prism distortion the only quadratic models
with the three properties mentioned above? Or do they belong to a
larger family of models from which we can select a better choice?
\item How can we combine these models or extend them while keeping all these
properties?
\item Is it necessary to sacrifice some of these properties in order to
obtain models with larger number of parameters?
\end{itemize}
In this work we intend to complement the physical approach to the
analysis of lens distortion models with a geometrical perspective.
To this purpose we formalize the relevant geometric properties of
the models and obtain those that comply with these properties. In
this way, we are in conditions to check to what extent the most employed
models enjoy these properties and propose extensions that preserve
them. 

The paper is organized as follows. In section \ref{sec:models} we
formalize the concept of lens distortion model and the main geometric
properties of interest. In section \ref{sec:Polynomial-models} we
study the basic properties of polynomial models introducing their
complex representation that will be essential in the later analysis.
Section \ref{sec:Linear-isotropic-models} includes the first result
of this work, which is the specification of all the possible polynomial
linear isotropic lens distortion models. Section \ref{sec:reflection-symmetric-models}
elaborates on this result, providing all the models that enjoy the
previous properties and at the same time are formed of functions with
reflection symmetry. Section \ref{sec:Geometry-of-some-models} analyzes
the properties of the most popular polynomial lens distortion models,
placing them in the framework introduced by the theoretical results
of the previous sections. Some extensions of these models are considered
in section \ref{sec:Experiments}, that also includes the corresponding
experiments. The conclusions are provided in section \ref{sec:Conclusions}.
An appendix at the end gathers the proofs of the theorems.

\section{Lens distortion models \label{sec:models}}

\subsection{Distortion functions\label{subsec:Distortion-functions}}

We will term \emph{lens distortion function} with \emph{distortion
center} $\mathbf{p}_{0}$ a smooth mapping $F:\mathbb{R}^{2}\to\mathbb{R}^{2}$
that keeps fixed $\mathbf{p}_{0}$ and has identity Jacobian $J(F)$
at this point. To simplify the formulation we will assume that $\mathbf{p}_{0}$
is at the origin of coordinates. This is not restrictive in most practical
situations, since the center of distortion is usually assumed to coincide
with the principal point of the projection. Then the distortion function
can be written as a mapping of the form 
\[
F(\mathbf{p})=\mathbf{p}+G(\mathbf{p})
\]
where $G(\mathbf{0})=\boldsymbol{0}$ and $JG(\mathbf{0})=\mathbf{0}$.
Function $G$ will be termed \emph{displacement} \emph{function}.
With this definition we are separating the linear and non-linear parts
of the imaging process, the linear part being associated to the intrinsic
parameter matrix. Two interesting analytical properties of lens distortion
functions are easy to check: 
\begin{itemize}
\item Each distortion function has a local inverse that is also of the same
form. 
\item The composition of two distortions functions is another function of
the same form. 
\end{itemize}
Some physical properties of the imaging system have a correspondence
with geometric properties of the displacement function. If the lens
has perfect rotational symmetry and the image plane is perfectly orthogonal
to the lens symmetry axis, the displacement function must be rotationally
invariant. Formally, if $R_{\theta}$ represents the planar rotation
of angle $\theta$, given by 
\begin{equation}
\mathbf{p}=(x,y)^{\top}\mapsto R_{\theta}(\mathbf{p})=\mathtt{R}_{\theta}\mathbf{p},\,\mathtt{R}_{\theta}=\begin{pmatrix}\cos\theta & -\sin\theta\\
\sin\theta & \cos\theta
\end{pmatrix},\label{eq:def-rotation}
\end{equation}
a displacement function $G$ is \emph{rotationally invariant} if it
satisfies 
\[
G=R_{-\theta}\circ G\circ R_{\theta}
\]
where $\circ$ denotes function composition.

Lack of parallelism between lens and image plane results in an image
formation system that is no longer rotationally symmetric, but is
symmetric with respect to the plane through the optical axis orthogonal
to both lens and image planes. Displacement functions corresponding
to this situation should exhibit \emph{reflection symmetry} with respect
to some line through the distortion center (symmetry axis). Formally,
if $T_{\mathbf{u}}$ is the reflection leaving invariant the line
through the origin with director vector $\mathbf{u}$, we have 
\[
G=T_{\mathbf{u}}\circ G\circ T_{\mathbf{u}}.
\]

The displacement function $G(x,y)$ of a lens distortion model can
be seen as a vector field on $\mathbb{R}^{2}$ that vanishes at the
origin. An orthogonal basis for such vector fields is given by $\mathbf{u}(x,y)=(x,y)^{\intercal}$,
$\mathbf{v}(x,y)=(-y,x)^{\intercal}$. Therefore, each displacement
function can be written univoquely as the sum of a \emph{radial} and
a \emph{tangential} displacement functions: 
\begin{equation}
\begin{pmatrix}x'\\
y'
\end{pmatrix}=\begin{pmatrix}x\\
y
\end{pmatrix}+\begin{pmatrix}x\\
y
\end{pmatrix}g_{r}(x,y)+\begin{pmatrix}-y\\
x
\end{pmatrix}g_{t}(x,y).\label{eq:rad-tan-decomposition}
\end{equation}

\subsection{Distortion models}

We define a \emph{lens distortion model }${\cal M}$ as a set of set
of displacement functions. A model will be termed \emph{linear} if
it is a vector space under the natural operations of sum and multiplication
by scalars. Linear models are of practical importance because they
greatly simplify the computational processes of obtainment of camera
parameters.

A model is \emph{isotropic} if it is invariant, as a set of functions,
with respect to coordinate rotations. It is natural to consider in
practice only models having this property because otherwise the characteristics
of the model would vary with a rotation of the data. Formally, if
$G$ is any function of the model ${\cal M}$, the model is \emph{isotropic}
if there is a $\tilde{G}\in{\cal M}$ such that 
\begin{equation}
\tilde{G}=R_{-\theta}\circ G\circ R_{\theta}.\label{eq:actionrotation}
\end{equation}

We will also pay special attention to those models including only
functions that are reflection symmetric with respect to some axis.

\section{Polynomial models\label{sec:Polynomial-models}}

\subsection{Polynomial lens displacement functions}

The $n$th-degree\emph{ polynomial lens distortion model} is the set
of displacement functions of the form 
\begin{equation}
\begin{pmatrix}\Delta x\\
\Delta y
\end{pmatrix}=\begin{pmatrix}X(x,y)\\
Y(x,y)
\end{pmatrix},\label{eq:real-pol-formulation}
\end{equation}
where $X$ and $Y$ are polynomials of degree $\leq n$ without linear
terms, so its Jacobian vanishes. We will also consider \emph{homogeneous}
$n$th-degree polynomial models in which $X$ and $Y$ are homogeneous
polynomials of degree $n$. 

For an arbitrary degree $n$ we define the vector mapping 
\begin{equation}
v_{n}(x,y)=(x^{n},x^{n-1}y,\ldots,y^{n})^{\intercal},\label{eq:v_n_definition}
\end{equation}
so that we can express homogeneous displacement functions as
\[
\begin{pmatrix}\Delta x\\
\Delta y
\end{pmatrix}=\begin{pmatrix}\boldsymbol{w}_{0}^{\intercal}\\
\boldsymbol{w}_{1}^{\intercal}
\end{pmatrix}v_{n}(x,y)=\mathtt{M}v_{n}(x,y),\,\mathbf{w}_{i}\in\mathbb{R}^{n+1}.
\]
General (i.e., non-homogeneous displacement functions) can be expressed
as sum of homogeneous displacement functions, and, consequently, can
be represented by sets of matrices.
\begin{example}
The simplest case is the quadratic model, corresponding to $n=2$,
for which the general and the homogeneous cases coincide. The displacement
functions are of the form: 
\begin{equation}
\begin{aligned}\Delta x & =a_{0}x^{2}+a_{1}xy+a_{2}y^{2}\\
\Delta y & =b_{0}x^{2}+b_{1}xy+b_{2}y^{2},\\
 & a_{i},b_{j}\in\mathbb{R},
\end{aligned}
\label{eq:quadratic-model-init}
\end{equation}
that can be expressed in matrix form as 
\begin{equation}
\begin{pmatrix}\Delta x\\
\Delta y
\end{pmatrix}=\begin{pmatrix}a_{0} & a_{1} & a_{2}\\
b_{0} & b_{1} & b_{2}
\end{pmatrix}\begin{pmatrix}x^{2}\\
xy\\
y^{2}
\end{pmatrix}.\label{eq:quadratic-model-1}
\end{equation}
\end{example}
A \emph{polynomial} \emph{radial displacement} is of the form 
\[
\begin{pmatrix}\Delta x\\
\Delta y
\end{pmatrix}=\begin{pmatrix}x\\
y
\end{pmatrix}p(x,y),
\]
where $p$ is a polynomial. As an example we have the well known $n$-coefficient
radial rotationally invariant (RRI) model, given by functions of the
form 
\begin{equation}
\begin{aligned}\begin{pmatrix}\Delta x\\
\Delta y
\end{pmatrix} & =\begin{pmatrix}x\\
y
\end{pmatrix}\left(\alpha_{1}r^{2}+\cdots+\alpha_{n}r^{2n}\right)\\
r^{2} & =x^{2}+y^{2}.
\end{aligned}
\label{eq:RRI-model}
\end{equation}
It is easy to check that all the polynomial radial distortions that
are invariant with respect to rotations are of this form.

We define analogously the \emph{polynomial tangential displacement}
functions as those of the form 
\[
\begin{pmatrix}\Delta x\\
\Delta y
\end{pmatrix}=\begin{pmatrix}-y\\
x
\end{pmatrix}q(x,y),
\]
where $q$ is a polynomial.

In the homogeneous case radial displacement functions can be expressed
as 
\begin{equation}
\begin{aligned}\begin{pmatrix}\Delta x\\
\Delta y
\end{pmatrix} & =\begin{pmatrix}x\\
y
\end{pmatrix}\mathbf{w}^{\top}v_{n-1}(x,y)\\
 & =\begin{pmatrix}w_{1} & \cdots & w_{n} & 0\\
0 & w_{1} & \cdots & w_{n}
\end{pmatrix}v_{n}(x,y),
\end{aligned}
\label{eq:hom-rad}
\end{equation}
and tangential distortion functions as 
\begin{equation}
\begin{aligned}\begin{pmatrix}\Delta x\\
\Delta y
\end{pmatrix} & =\begin{pmatrix}-y\\
x
\end{pmatrix}\mathbf{w}^{\top}v_{n-1}(x,y)\\
 & =\begin{pmatrix}0 & -w_{1} & \cdots & -w_{n}\\
w_{1} & \cdots & w_{n} & 0
\end{pmatrix}v_{n}(x,y).
\end{aligned}
\label{eq:hom-tan}
\end{equation}

Therefore radial and tangential displacement functions constitute
linear subspaces of dimension $n$ of the matrix space $\mathbb{R}^{2\times(n+1)}$,
that intersect trivially. Since the dimension of the matrix space
is $2(n+1)>2n$, the functions $g_{r}$ and $g_{t}$ in the decomposition
\eqref{eq:rad-tan-decomposition} are not in general polynomial for
a polynomial displacement function. So we have the following proposition.
\begin{proposition}
\label{th:radial-tangential-spaces}The sets of $n$th-degree homogeneous
radial or tangential displacements constitute isotropic subspaces
of dimension $n$ of the matrix space $\mathbb{R}^{2\times(n+1)}$,
that intersect trivially.
\end{proposition}
\begin{example}
In the quadratic case the radial displacements are those of the form

\begin{equation}
\begin{aligned}\begin{pmatrix}x\\
y
\end{pmatrix}\left(t_{1}x+t_{2}y\right)=\begin{pmatrix}t_{1} & t_{2} & 0\\
0 & t_{1} & t_{2}
\end{pmatrix}\begin{pmatrix}x^{2}\\
xy\\
y^{2}
\end{pmatrix},\end{aligned}
\label{eq:quad-radial}
\end{equation}
and the tangential displacements are those of the form

\begin{equation}
\begin{aligned}\begin{pmatrix}-y\\
x
\end{pmatrix}\left(u_{1}x+u_{2}y\right) & =\begin{pmatrix}0 & -u_{1} & -u_{2}\\
u_{1} & u_{2} & 0
\end{pmatrix}\begin{pmatrix}x^{2}\\
xy\\
y^{2}
\end{pmatrix}.\end{aligned}
\label{eq:quad-tangential}
\end{equation}
The direct sum of the corresponding linear models is a vector subspace
of dimension four of $\mathbb{R}^{2\times3}$, with which we can identify
the set of quadratic distortion functions. Any quadratic displacement
function outside this four-dimensional subspace has non-polynomial
radial or tangential components. 
\end{example}

\subsection{Complex polynomial formulation of displacement functions}

Polynomial displacement functions \eqref{eq:real-pol-formulation}
can be expressed equivalently as a single complex polynomial in the
complex variables $z$ and $\bar{z}$, 

\begin{equation}
f(z,\bar{z})=\Delta z=\sum_{(k,l)\in I}^{n}\gamma_{kl}z^{k}\bar{z}^{l},\,\gamma_{kl}\in\mathbb{C},\label{eq:complex-pol-formulation}
\end{equation}
where $I$ is any finite set of index pairs $(k,l)$ such that $k\geq0$,
$l\geq0$, $k+l\geq2$. These polynomials have not been so far, to
the authors knowledge, employed to express lens distortion functions,
and we will see that they facilitate enormously the geometrical analysis
of models.

The real polynomial \eqref{eq:real-pol-formulation} and the complex
polynomial formulations \eqref{eq:complex-pol-formulation} are indeed
equivalent, since, if we write $P(x,y)=X(x,y)+iY(x,y)$, we have that
\[
P(x,y)=P\left(\frac{1}{2}(z+\overline{z}),\frac{1}{2i}(z-\overline{z})\right)=f(z,\overline{z}).
\]
Conversely, since $z=x+iy$, we recover $P=X+iY$ from $f$. 
\begin{example}
In the quadratic case a general complex polynomial is given by 
\[
\Delta z=\gamma_{20}z^{2}+\gamma_{11}z\bar{z}+\gamma_{02}\bar{z}^{2}.
\]
Let us write $\gamma_{kl}=\alpha_{kl}+i\beta_{kl}$. The corresponding
real polynomial expression will be of the form 
\[
\Delta\mathbf{p}=\begin{pmatrix}a_{0} & a_{1} & a_{2}\\
b_{0} & b_{1} & b_{2}
\end{pmatrix}\begin{pmatrix}x^{2}\\
xy\\
y^{2}
\end{pmatrix}.
\]
If we denote $\mathbf{a}=(a_{0},a_{1},a_{2})^{\intercal}$, $\mathbf{b}=(b_{0},b_{1},b_{2})^{\intercal}$,
$\boldsymbol{\alpha}=(\alpha_{20},\alpha_{11},\alpha_{02})^{\intercal}$,
$\boldsymbol{\beta}=(\beta_{20},\beta_{11},\beta_{02})^{\intercal}$
and $\mathbf{c}=\mathbf{a}+i\mathbf{b}$, $\boldsymbol{\gamma}=\boldsymbol{\alpha}+i\boldsymbol{\beta}$,
it is easy to check that the correspondence between both sets of parameters
is given by
\[
\mathbf{c}=\mathtt{C}\boldsymbol{\gamma},
\]
where 
\[
\mathtt{C}=\left(\begin{array}{rrr}
1 & 1 & 1\\
2i & 0 & -2i\\
-1 & 1 & -1
\end{array}\right).
\]
The matrix $\mathtt{C}$ is invertible as a consequence of the equivalence
between both kinds of parameterizations. 
\end{example}
Radial and tangential displacement functions are also easily expressed
in complex polynomial notation. Since $z$ corresponds to the radial
vector $(x,y)$ and $iz$ to the tangential vector $(-y,x)$, radial
and tangential displacements are given respectively by expressions
of the form
\[
zp(z,\bar{z}),\,izq(z,\bar{z}),
\]
where $p(z,\bar{z)}$ and $q(z,\bar{z})$ are \emph{real-valued complex
polynomials}, i.e., such that for any $z\in\mathbb{C}$ their evaluation
is real. It is easy to check that this is equivalent to having coefficients
satisfying $\gamma_{kl}=\bar{\gamma}_{lk}$. 

Therefore the complex polynomials that are multiples of $z$ represent
displacement functions that lie in the space generated by radial and
tangential displacement functions. The only monomials that do not
lie in this space are those of the form $\bar{z}^{n}$, thus providing
a natural complement of that space (see proposition \ref{th:radial-tangential-spaces}).

\section{Linear isotropic models\label{sec:Linear-isotropic-models}}

In this section we aim at obtaining the polynomial models that enjoy
at the same time the properties of being linear and rotationally invariant.
To this purpose we will make use of the theory of group representations.

\subsection{Group representations on polynomial spaces\label{subsec:group-representations}}

Given a group $G$, a representation of $G$ on a vector space $V$
is a group homomorphism 
\[
\rho:G\longrightarrow\mathrm{Aut}(V),
\]
where $\mathrm{Aut}(V)$ stands for the group of automorphisms of
$V$, i.e., the set of invertible linear mappings $f:V\to V$. Hence,
a representation is just a group action on the vector space $V$ such
that the transformations defined by the elements of $G$ are linear
mappings $V\to V$.

As an example that will be useful for our purposes, let us consider
the group $G=SO(2)$ of plane rotations and the vector space $V=\mathcal{H}^{n}$
of homogeneous polynomials $P:\mathbb{R}^{2}\to\mathbb{R}$ of degree
$n$ in the variables $(x,y)$. The group representation 
\[
\rho:SO(2)\longrightarrow\mathrm{Aut}({\cal H}^{n})
\]
is simply given by $\rho(\mathtt{R}_{\theta})(P)=P'$ where 
\[
P'(\mathbf{p})=P(\mathtt{R}_{\theta}\mathbf{p}),
\]
where $\mathbf{p}=(x,y)^{\intercal}$. It is immediate to check that
$\rho(\mathtt{R}_{\theta})$ is a linear mapping whose inverse is
$\rho(\mathtt{R}_{-\theta})$.

Since $\rho(\mathtt{R}_{\theta})$ is an automorphism of \foreignlanguage{english}{${\cal H}^{n}$},
the elements of the basis of ${\cal H}^{n}$ given by the components
of $v_{n}(\mathbf{p})$ (defined in \eqref{eq:v_n_definition}) are
transformed into the basis 
\begin{align*}
\left(\rho(\mathtt{R}_{\theta})(x^{n}),\rho(\mathtt{R}_{\theta})(x^{n-1}y),\ldots,\rho(\mathtt{R}_{\theta})(y^{n})\right)^{\intercal}\\
=\rho(\mathtt{R}_{\theta})(v_{n}(\mathbf{p}))=v_{n}(\mathtt{R}_{\theta}\mathbf{p}),
\end{align*}
 and so there exists a regular matrix $V_{n}(\mathtt{R}_{\theta})$
of order $n+1$ such that

\begin{equation}
v_{n}(\mathtt{R}_{\theta}\mathbf{p})=\mathtt{V}_{n}(\mathtt{R}_{\theta})v_{n}(\mathbf{p}).\label{eq:def-Vn}
\end{equation}
For instance, for $n=2$ we have

\[
\mathtt{V}_{2}(\mathtt{R}_{\theta})=\begin{pmatrix}\cos^{2}\theta & -\sin2\theta & \sin^{2}\theta\\
\frac{1}{2}\sin2\theta & \cos2\theta & -\frac{1}{2}\sin2\theta\\
\sin\theta^{2} & \sin2\theta & \cos^{2}\theta
\end{pmatrix}.
\]

A vector subspace $W\subset V$ is called $G$-invariant if $\rho(g)(W)\subset W$
for every $g\in G$. A representation $\rho:G\to\mathrm{Aut}(V)$
is said to be irreducible if there exist no $G$-invariant subspace
but the trivial ones, i.e., the null-subspace and $V$ itself. 

An important property of compact groups as $SO(2)$ is that any representation
is completely reducible, i.e., the associated vector space can be
decomposed as $V=V_{1}\oplus\cdots\oplus V_{N}$, the restriction
of the representation $\rho$ to any $V_{i}$ being an irreducible
representation \cite{Vinberg}.

\subsection{Polynomial displacements and geometric transformations}

The set of homogeneous displacement functions of degree $n$ $\mathbf{P}:\mathbb{R}^{2}\to\mathbb{R}^{2}$,
$\mathbf{P}(x,y)=(X(x,y),Y(x,y))$ is a vector space ${\cal V}^{n}$
in which the plane rotation group $SO(2)$ acts according to equation
(\ref{eq:actionrotation}). Specifically, a rotation transforms the
mapping $\mathbf{P}$ into the mapping $\mathbf{P}'$ given by 
\[
\mathbf{P}'(\mathbf{x})=\mathtt{R}_{\theta}^{\intercal}\mathbf{P}\left(\mathtt{R}_{\theta}\mathbf{x}\right),
\]
where $\mathbf{x}=(x,y)^{\intercal}$ and $\mathtt{R}_{\theta}$ is
defined in \eqref{eq:def-rotation}.

Let us consider in more detail the homogeneous case. The displacement
function is then given by the equation 
\begin{equation}
\Delta\mathbf{p}=\mathtt{M}v_{n}(\mathbf{p}),\,\Delta\mathbf{p}\equiv\begin{pmatrix}\Delta x\\
\Delta y
\end{pmatrix},\,\mathbf{p}\equiv\begin{pmatrix}x\\
y
\end{pmatrix},\label{eq:hom-dist-fun}
\end{equation}
where $\mathtt{M}$ is a $2\times(n+1)$ matrix. In order to see how
matrix $\mathtt{M}$ in \eqref{eq:hom-dist-fun} changes with coordinate
rotation we substitute in this equation 
\[
\mathbf{p}=\mathtt{R}\bar{\mathbf{p}},\,\Delta\mathbf{p}=\mathtt{R}\Delta\bar{\mathbf{p}},
\]
obtaining 
\[
\begin{aligned}\Delta\bar{\mathbf{p}} & =\mathtt{R}^{\top}\mathtt{M}v_{n}\left(\mathtt{R}\bar{\mathbf{p}}\right)\\
 & =\mathtt{R}^{\top}\mathtt{M}\mathtt{V}_{n}\left(\mathtt{R}\right)v_{n}\left(\bar{\mathbf{p}}\right)\\
 & =\bar{\mathtt{M}}v_{n}\left(\bar{\mathbf{p}}\right),
\end{aligned}
\]
where
\begin{equation}
\bar{\mathtt{M}}=\mathtt{R}^{\top}\mathtt{M}\mathtt{V}_{n}(\mathtt{R}).\label{eq:hom-coord-rot}
\end{equation}

Thus a homogeneous distortion function transforms itself under the
action of a coordinate rotation into another one given by the previous
formula. And, in particular, we have that polynomial models, homogeneous
or not, are isotropic.

The complex function formulation \eqref{eq:complex-pol-formulation}
allows for an easier treatment of coordinate rotation. Using complex
numbers, a coordinate rotation of angle $\theta$ can be written as
\[
z=e^{i\theta}w,\,\Delta z=e^{i\theta}\Delta w.
\]
Let us see how these changes of variables induce a transformation
in the complex polynomial. We have 
\[
e^{i\theta}\Delta w=\sum_{(k,l)\in I}\gamma_{kl}e^{i\theta(k-l)}w^{k}\bar{w}^{l},
\]
so that the new polynomial is 
\begin{equation}
\Delta w=\sum_{(k,l)\in I}\gamma_{kl}e^{i\theta(k-l-1)}w^{k}\bar{w}^{l}.\label{eq:cp-pol-rot}
\end{equation}

In the case of monomials, the corresponding transformation is
\begin{equation}
z^{k}\bar{z}^{l}\mapsto e^{i\theta(k-l-1)}w^{k}\bar{w}^{l}.\label{eq:cp-pol-rot-monomials}
\end{equation}
We will call the number $m=k-l-1$ the \emph{winding number} of the
monomial. Table \ref{table:monomials-by-exponent} shows a classification
of the monomials of degrees from two to five according to their associated
winding number.
\begin{example}
For degree two a coordinate rotation transforms the coefficients according
to 
\begin{equation}
(\gamma_{20},\gamma_{11},\gamma_{02})\mapsto(e^{i\theta}\gamma_{20},e^{-i\theta}\gamma_{11},e^{-3i\theta}\gamma_{02}).\label{eq:cp-pol-rot-degree-2}
\end{equation}
\end{example}
\begin{table*}
\centering{}%
\begin{tabular}{|c|c|c|c|c|c|c|c|c|c|c|c|}
\hline 
$m$ & $-6$ & $-5$ & $-4$ & $-3$ & $-2$ & $-1$ & 0 & 1 & 2 & 3 & 4\tabularnewline
\hline 
\hline 
 &  &  &  & $\bar{z}^{2}$ &  & $z\bar{z}$ &  & $z^{2}$ &  &  & \tabularnewline
\hline 
 &  &  & $\bar{z}^{3}$ &  & $z\bar{z}^{2}$ &  & $z^{2}\bar{z}$ &  & $z^{3}$ &  & \tabularnewline
\hline 
 &  & $\bar{z}^{4}$ &  & $z\bar{z}^{3}$ &  & $z^{2}\bar{z}^{2}$ &  & $z^{3}\bar{z}$ &  & $z^{4}$ & \tabularnewline
\hline 
 & $\bar{z}^{5}$ &  & $z\bar{z}^{4}$ &  & $z^{2}\bar{z}^{3}$ &  & $z^{3}\bar{z}^{2}$ &  & $z^{4}\bar{z}$ &  & $z^{5}$\tabularnewline
\hline 
\end{tabular}\caption{Classification of monomials up to degree five by their winding number.}
\label{table:monomials-by-exponent}
\end{table*}

\subsection{Rotation-invariant distortion functions}

We will call \emph{invariant monomials} those of zero winding number,
i.e., those that are invariant with respect to coordinate rotations
\eqref{eq:cp-pol-rot-monomials}. They are of the form
\begin{equation}
z^{k+1}\bar{z}^{k},\,k>0,\label{eq:invariant-monomials}
\end{equation}
and therefore there are no invariant monomials of even degree. The
displacement functions that do not change under coordinate rotations
are those given by complex linear combinations of invariant monomials. 

We can write the term corresponding to an invariant monomial $\gamma z^{k}\bar{z}^{k+1}$
as the sum of a radial and a tangential term as 
\[
\gamma z^{k}\bar{z}^{k+1}=z\left(az^{k}\bar{z}^{k}\right)+(iz)\left(bz^{k}\bar{z}^{k}\right),
\]
with $\gamma$ being $a+ib$.

\begin{figure}
\begin{centering}
\includegraphics[width=0.85\columnwidth]{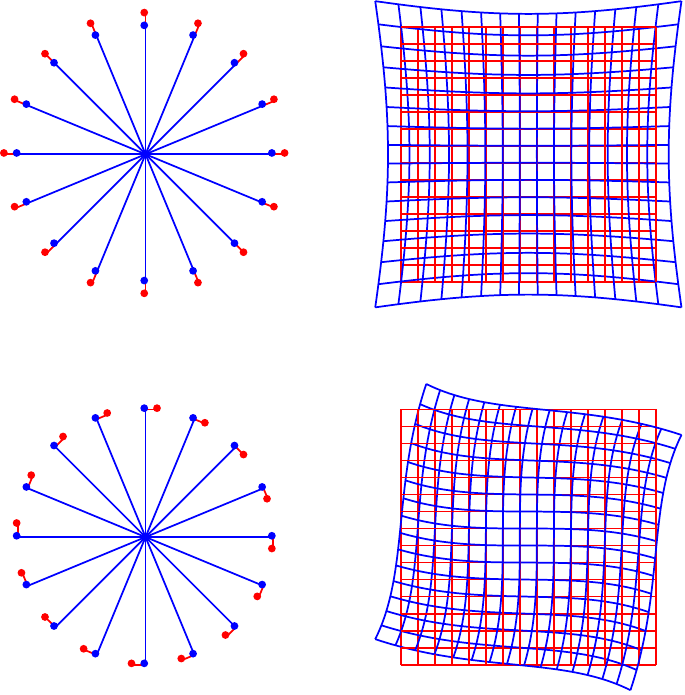} 
\par\end{centering}
\caption{\label{fig:cub-rot-inv}Action on a circle and on a grid of the rotationally
invariant cubic distortions corresponding to matrices \eqref{eq:cubic-RRI-matrices}.
Top: radial invariant distortion, bottom: tangential invariant distortion.}
\end{figure}

In the case of degree three, the radial and tangential terms correspond
respectively to the matrices
\begin{equation}
\begin{pmatrix}1 & 0 & 1 & 0\\
0 & 1 & 0 & 1
\end{pmatrix}\,\text{and }\begin{pmatrix}0 & 1 & 0 & 1\\
-1 & 0 & -1 & 0
\end{pmatrix}.\label{eq:cubic-RRI-matrices}
\end{equation}
The first one corresponds to the cubic (one-parameter) invariant radial
distortion of equation \eqref{eq:RRI-model} and the other one to
invariant tangential distortion. Figure \ref{fig:cub-rot-inv} shows
the action of the corresponding distortion functions on points of
a circle and on a grid. 

\subsection{Linear isotropic models\label{subsec:Linear-isotropic-models}}

In this subsection we obtain all the linear isotropic polynomial models
of functions of a given maximum degree. In the language of group representations,
these are the invariant subspaces of the representation of the planar
rotation group on the vector space of displacement functions. As we
mentioned in section \ref{subsec:group-representations}, these invariant
subspaces are direct sum of irreducible invariant subspaces. Therefore
the problem is that of finding these irreducible subspaces.

Some notation will be useful in the sequel. We will denote by $\mathcal{P}^{(n)}$
the complex vector space of polynomials $f(z,\bar{z})$ spanned by
the monomials $z^{k}\bar{z}^{l}$ of degree $k+l\in\left\{ 2,\ldots,n\right\} $,
by $\mathcal{P}_{m}^{(n)}$ the subspace of $\mathcal{P}^{(n)}$ generated
by the monomials with winding number $m$ and ${\cal W}^{(n)}$ the
subspace generated by all the monomials with winding number $m\neq0$,
i.e., the non-invariant monomials. Therefore we have
\begin{align*}
\mathcal{P}^{(n)} & ={\cal P}_{0}^{(n)}\oplus{\cal W}^{(n)},\\
{\cal W}^{(n)} & =\bigoplus_{m\neq0}\mathcal{P}_{m}^{(n)}.
\end{align*}

Let us denote by $\mathbb{P}_{\mathbb{C}}^{1}=\mathbb{C}^{2}\setminus\left\{ (0,0)\right\} /\mathbb{C}^{*}$
the complex projective line. Its points are equivalence classes 
\[
\left[\left(\mu,\nu\right)\right]=\left\{ \left(\gamma\mu,\gamma\nu\right):\gamma\in\mathbb{C}^{*}\right\} .
\]
We will denote $\left[\left(\mu,\nu\right)\right]=(\mu:\nu)$. Analogously,
the real projective line $\mathbb{P}_{\mathbb{R}}^{1}=\mathbb{R}^{2}\setminus\left\{ (0,0)\right\} /\mathbb{R}^{*}=\mathbb{C}^{*}/\mathbb{R}^{*}$
and its points will be denoted as $[\mu]$ for $\mu\in\mathbb{C^{*}}$.

Since $\mathcal{P}^{(n)}=\mathcal{P}_{0}^{(n)}\oplus{\cal W}^{(n)}$
and the elements of $\mathcal{P}_{0}^{(n)}$ are kept fixed by the
representation, we just have to obtain the irreducible subspaces of
${\cal W}^{(n)}$. Albeit the set $\mathcal{P}^{(n)}$ has a natural
structure of complex vector space, we are interested in $\mathcal{P}^{(n)}$
as a real vector space, since we are identifying it with pairs $(P(x,y),Q(x,y))$
of polynomials in two real variables. We will denote by $\mathcal{P}_{\mathbb{R}}^{(n)}$
this real vector space.
\begin{theorem}
\label{th:irreducible-real-subspaces-all} The irreducible real subspaces
of the representation $\rho:SO(2)\to\mathrm{Aut}(\mathcal{P}^{(n)})$
are the one-dimen\-sional real subspaces of $\mathcal{P}_{0}^{(n)}$
together with the bidimensional subspaces of the form
\begin{equation}
\mathcal{M}_{m}^{(n)}\left[f,g\right]=\left\{ \gamma f(z,\bar{z})+\bar{\gamma}g(z,\bar{z})\colon\gamma\in\mathbb{C}\right\} ,\label{eq:Mm_uv_spaces}
\end{equation}
where $f\in\mathcal{P}_{m}^{(n)}$, $g\in\mathcal{P}_{-m}^{(n)}$.
\end{theorem}
\begin{svmultproof}
Consider the basis of $\mathcal{P}_{\mathbb{R}}^{(n)}$ 
\[
\mathcal{B}=\left\{ z^{k}\overline{z}^{l},iz^{k}\overline{z}^{l}\right\} _{k,l\geq0,\,2\leq k+l\leq n},
\]
where we suppose that the monomials are ordered by their winding number
$m=k-l-1$. Since 
\[
\rho(e^{i\theta})(z^{k}\overline{z}^{l})=e^{im\theta}z^{k}\overline{z}^{l},
\]
the matrix $\mathtt{M}$ of the automorphism $\rho(e^{i\theta})$
with respect to $\mathcal{B}$ is built with diagonal blocks
\[
\mathtt{M}_{m}=\begin{pmatrix}\cos m\theta & -\sin m\theta\\
\sin m\theta & \cos m\theta
\end{pmatrix}.
\]
An irreducible invariant real subspace $W$ of $\mathtt{M}$ must
be associated to a pair of complex conjugate eigenvalues, which necessarily
are of the form $e^{im\theta},e^{-im\theta}$. Therefore $W$ must
be an irreducible invariant subspace of 
\[
\mathcal{P}_{m}^{(n)}\oplus\mathcal{P}_{-m}^{(n)}.
\]
Such subspaces are obtained in lemma \ref{lemma:real-invariant-subspaces}
and are of the form $\left\{ \gamma f(z,\bar{z})+\bar{\gamma}g(z,\bar{z})\colon\gamma\in\mathbb{C}\right\} $,
$f\in\mathcal{P}_{m}^{(n)}$, $g\in\mathcal{P}_{-m}^{(n)}$, as stated. 
\end{svmultproof}

\begin{remark}
\label{rem:irreducible-parametrization}Observe that $\mathcal{M}_{m}^{(n)}\left[f,g\right]$
and $\mathcal{M}_{m}^{(n)}\left[\tilde{f},\tilde{g}\right]$ are the
same space if and only if $\tilde{f}=\alpha f$, $\tilde{g}=\bar{\alpha}g$
for some $\alpha\in\mathbb{C}^{*}$. Otherwise the spaces have trivial
intersection.
\end{remark}
\begin{example}
\label{example:irred-dim-2}In degree $n=2$ we have only three monomials,
each of them with a different winding number: $z^{2}$ $(m=1)$, $z\bar{z}$
($m=-1$) and $\bar{z}^{2}$ ($m=-3$). Therefore there are no invariant
monomials. Thus a generic polynomial of ${\cal P}_{1}^{(2)}$ is of
the form $f=\mu z^{2}$, $\mu\in\mathbb{C}$, and a generic polynomial
of ${\cal P}_{-1}^{(2)}$ is of the form $g=\bar{\nu}z\bar{z}$. Therefore,
we can parameterize the set of irreducible invariant subspaces $\mathcal{M}_{m}^{(n)}\left[f,g\right]$
by the pair of coefficients $(\mu,\nu)$, and since, by remark \ref{rem:irreducible-parametrization},
$(\mu,\nu)$ and $(\alpha\mu,\alpha\nu)$ produce the same space,
we have that the irreducible subspaces of ${\cal P}_{1}^{(2)}\oplus{\cal P}_{-1}^{(2)}$
can be adequately parameterized by the projective points $(\mu:\nu)\in\mathbb{P}_{\mathbb{C}}^{1}$.
These subspaces are thus given by

\begin{align}
\mathcal{M}_{1}^{(2)}(\mu:\nu) & =\left\{ \gamma\mu z^{2}+\bar{\gamma}\bar{\nu}z\bar{z}\colon\gamma\in\mathbb{C}\right\} ,\,(\mu:\nu)\in\mathbb{P}_{\mathbb{C}}^{1}.\label{eq:irreducible-degree-2}
\end{align}
Observe that 
\[
\mathcal{M}_{1}^{(2)}(1:1)=\left\{ z\left(\gamma z+\bar{\gamma}\bar{z}\right)\colon\gamma\in\mathbb{C}\right\} ,
\]
with $\gamma\bar{z}+\bar{\gamma}z$ being real-valued, is the space
of radial displacements and 
\[
\mathcal{M}_{1}^{(2)}(1:-1)=\left\{ z\left(\gamma z-\bar{\gamma}\bar{z}\right)\colon\gamma\in\mathbb{C}\right\} ,
\]
is the space of tangential displacements, as $\gamma z-\bar{\gamma}\bar{z}$
takes only pure imaginary values. Since different irreducible subspaces
intersect trivially, we have that the direct sum of any two different
subspaces of the form \eqref{eq:irreducible-degree-2} is the whole
four-dimensional space 
\begin{align}
{\cal P}_{1}^{(2)}\oplus{\cal P}_{-1}^{(2)} & =\left\{ \gamma_{1}z^{2}+\gamma_{2}z\bar{z}\colon\gamma_{1},\gamma_{2}\in\mathbb{C}\right\} \label{eq:quad-rad-tan}\\
 & =\mathcal{M}_{1}^{(2)}(1:1)\oplus\mathcal{M}_{1}^{(2)}(1:-1).\nonumber 
\end{align}
In section \ref{sec:Geometry-of-some-models} we will see another
interesting decomposition of this space (see equation \eqref{eq:another-decomposition}).

In the case of winding number $m=-3$ the subspace generated by the
only associated monomial,
\[
{\cal P}_{-3}^{(2)}=\left\{ \gamma\bar{z}^{2}\colon\gamma\in\mathbb{C}\right\} 
\]
already coincides with the irreducible invariant subspace $\mathcal{M}_{3}^{(2)}\left[\bar{z}^{2},0\right]$.
\end{example}

\section{Reflection-symmetric distortion functions \label{sec:reflection-symmetric-models}}

As we have mentioned before, distortion functions that have reflection
symmetry with respect to some axis are important in order to model
some optical phenomenons. In this section we obtain all the polynomial
models that enjoy at the same time the three properties of being linear,
isotropic, and being formed by functions with reflection symmetry.
We will see that this triple requirement happens to limit severely
the dimensionality of the possible models, thus pointing towards the
need of relaxing some of the constraints in order to gain flexibility.

\subsection{Equations and parameterizations of the variety}

The following theorem describes the polynomial displacement functions
with reflection symmetry.
\begin{proposition}
A polynomial displacement function 
\[
f(z,\bar{z})=\sum_{(k,l)\in I}\gamma_{kl}z^{k}\bar{z}^{l}
\]
 is reflection-symmetric with respect to the axis $\left\langle e^{i\theta}\right\rangle =\left\{ ae^{i\theta}:a\in\mathbb{R}\right\} $
if and only if it satisfies 
\[
e^{2i\theta}\overline{f(z,\bar{z})}=f(e^{2i\theta}\bar{z},e^{-2i\theta}z),
\]
which is equivalent to have coefficients of the form
\begin{align}
\gamma_{kl} & =a_{kl}e^{im\theta},\nonumber \\
 & a_{kl},\theta\in\mathbb{R},\,m=k-l-1,\label{eq:ref-sym-param}
\end{align}
 and therefore the coefficients satisfy the relation
\begin{equation}
\Im\left[\gamma_{kl}^{m'}\bar{\gamma}_{k'l'}^{m}\right]=0.\label{eq:rsf-coeff-equations}
\end{equation}
\end{proposition}
\begin{svmultproof}
A reflection with respect to the axis $\left\langle e^{i\theta}\right\rangle =\left\{ ae^{i\theta}:a\in\mathbb{R}\right\} $
is expressed in terms of complex numbers by the mapping
\[
z\mapsto e^{2i\theta}\bar{z}.
\]

Therefore a displacement 
\[
\Delta z=f(z,\bar{z})
\]
 is reflection-symmetric with respect to this axis if
\[
e^{2i\theta}\overline{\Delta z}=f(e^{2i\theta}\bar{z},e^{-2i\theta}z),
\]
i.e., if
\[
e^{2i\theta}\overline{f(z,\bar{z})}=f(e^{2i\theta}\bar{z},e^{-2i\theta}z).
\]
A straightforward computation shows that this is equivalent to have
coefficients satisfying
\begin{equation}
\gamma_{kl}=e^{-2i\theta m}\overline{\gamma}_{kl},\,m=k-l-1.\label{eq:ref-sym-gamma-cond}
\end{equation}
Writing $\gamma_{kl}=\rho_{kl}e^{i\phi_{kl}}$, with $\rho_{kl}\geq0$,
the equation above implies 
\[
e^{2i\phi_{kl}}=e^{-2i\theta m},
\]
i.e.,
\begin{align*}
2\phi_{kl} & =-2\theta m+2k\pi,k\in\mathbb{Z}\\
\Leftrightarrow\phi_{kl} & =-\theta m+k\pi\\
\Leftrightarrow\gamma_{kl} & =\rho_{kl}e^{-i\theta m}e^{ik\pi}=\pm\rho_{kl}e^{-i\theta m}.
\end{align*}

From \eqref{eq:ref-sym-gamma-cond}, for $(k,l)\neq(k',l')$, denoting
$m'=k'-l'-1$, we must have
\begin{equation}
\left(\frac{\gamma_{kl}}{\bar{\gamma}_{kl}}\right)^{m'}=\left(\frac{\gamma_{k'l'}}{\bar{\gamma}_{k'l'}}\right)^{m}.\label{eq:rsf-general-equations}
\end{equation}
i.e.,
\[
\gamma_{kl}^{m'}\bar{\gamma}_{k'l'}^{m}=\bar{\gamma}_{kl}^{m'}\gamma_{k'l'}^{m}
\]
or equivalently
\[
\Im\left[\gamma_{kl}^{m'}\bar{\gamma}_{k'l'}^{m}\right]=0.
\]
\end{svmultproof}

\begin{figure}
\begin{centering}
\includegraphics[width=0.85\columnwidth]{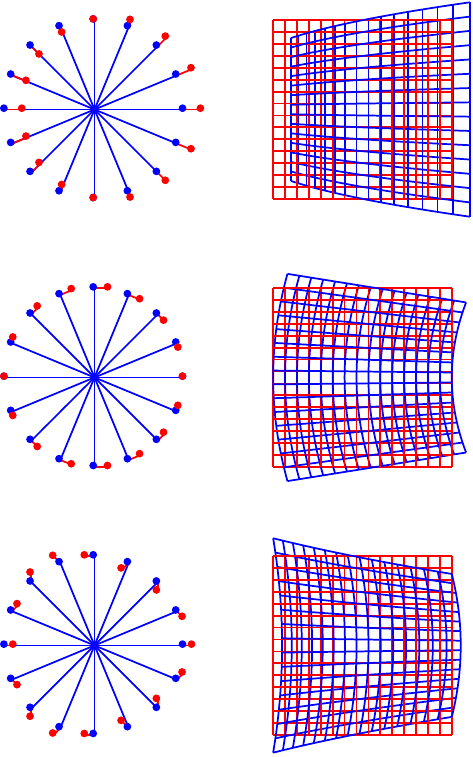} 
\par\end{centering}
\caption{\label{fig:quad-sym-basis}Quadratic distortions given by each of
the matrices in \eqref{eq:quad-sym-functions}, ordered from top to
bottom and symmetric with respect to the horizontal axis. Action on
points a circle and on a grid.}
\end{figure}

\begin{figure}
\begin{centering}
\includegraphics[width=0.85\columnwidth]{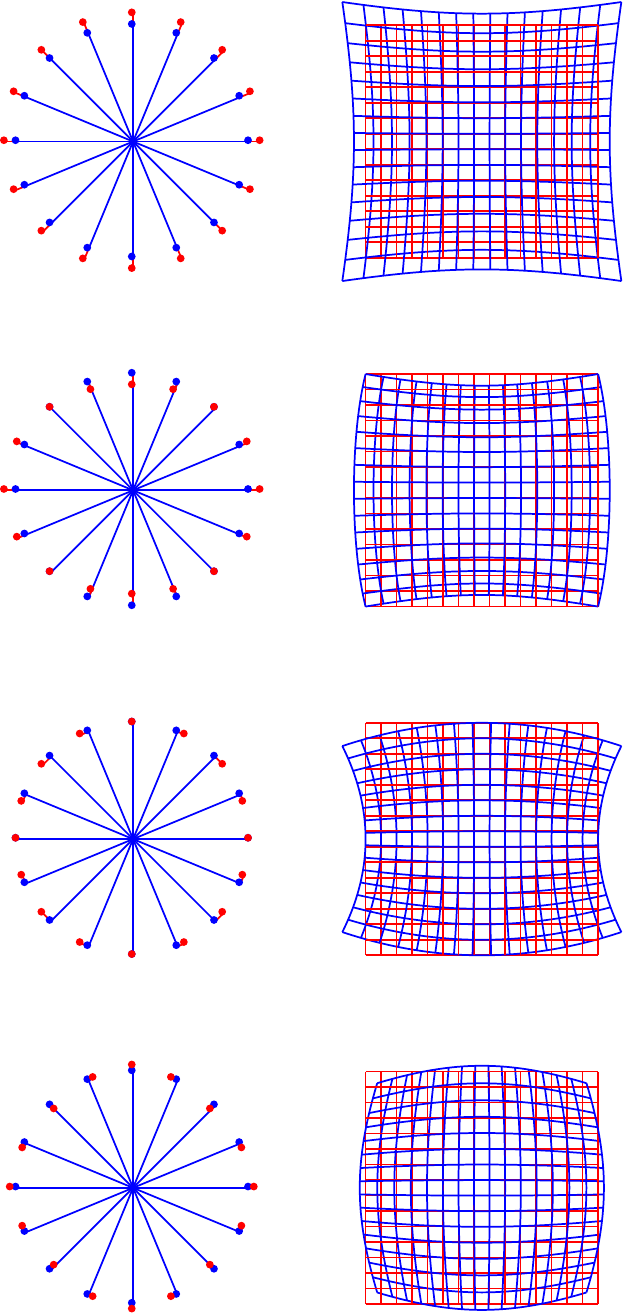} 
\par\end{centering}
\caption{\label{fig:cub-sym-basis} Cubic distortions given by each of the
matrices in \eqref{eq:cubic-sym-functions}, ordered from top to bottom
and symmetric with respect to the horizontal axis. Action on points
a circle and on a grid.}
\end{figure}

\begin{remark}
The equations \eqref{eq:rsf-coeff-equations} are sufficient conditions
if there exists a monomial with winding number $m=1$, as it is easy
to check. However, in the general case they are not sufficient conditions
as the polynomial 
\[
f(z,\bar{z})=z^{3}+iz\bar{z}^{2}
\]
shows.
\end{remark}
\begin{remark}
In particular, for the invariant monomials ($m=0$) this implies
\[
\hat{\gamma}_{kl}=a_{kl}\in\mathbb{R}.
\]
\end{remark}
\begin{example}
For degree two, the functions symmetric with respect to the horizontal
axis are 
\[
f(z,\bar{z})=a_{0}z^{2}+a_{1}z\bar{z}+a_{2}\bar{z}^{2},\,a_{i}\in\mathbb{R},
\]
and after coordinate rotation we obtain
\begin{equation}
\hat{f}(z,\bar{z})=a_{0}e^{i\theta}z^{2}+a_{1}e^{-i\theta}z\bar{z}+a_{2}e^{-3i\theta}\bar{z}^{2}.\label{eq:quad-sym-complex}
\end{equation}
Let us see that the first two terms can be written as the sum of a
radial term and a tangential term. Writing $a=a_{0}+a_{1},\,b=a_{0}-a_{1}$,
we have
\[
a_{0}e^{i\theta}z^{2}+a_{1}e^{-i\theta}z\bar{z}=az\frac{1}{2}(e^{i\theta}z+e^{-i\theta}\bar{z})+biz\frac{1}{2i}(e^{i\theta}z-e^{-i\theta}\bar{z}),
\]
so that in real polynomial form the first two terms of $\hat{f}(z,\bar{z})$
are 
\[
a\begin{pmatrix}x\\
y
\end{pmatrix}\left(x\cos\theta-y\sin\theta\right)+b\begin{pmatrix}-y\\
x
\end{pmatrix}\left(x\sin\theta+y\cos\theta\right),
\]
and in real matrix form, including the three terms, we obtain 
\begin{equation}
\begin{aligned}a\begin{pmatrix}\cos\theta & -\sin\theta & 0\\
0 & \cos\theta & -\sin\theta
\end{pmatrix}\\
+b\begin{pmatrix}0 & \sin\theta & \cos\theta\\
-\sin\theta & -\cos\theta & 0
\end{pmatrix}\\
+c\begin{pmatrix}\cos3\theta & -2\sin3\theta & -\cos3\theta\\
-\sin3\theta & -2\cos3\theta & \sin3\theta
\end{pmatrix}.
\end{aligned}
\label{eq:quad-sym-functions}
\end{equation}
Figure \ref{fig:quad-sym-basis} shows the action of each of these
terms on points on a circle and on a grid oriented according to the
symmetry axis.

If we consider functions of degree $n=3$ an analogous process leads
to the parameterization 
\begin{equation}
\begin{aligned}d\begin{pmatrix}1 & 0 & 1 & 0\\
0 & 1 & 0 & 1
\end{pmatrix}\\
+e\begin{pmatrix}\cos2\theta & -2\sin2\theta & -\cos2\theta & 0\\
0 & \cos2\theta & -2\sin2\theta & -\cos2\theta
\end{pmatrix}\\
+f\begin{pmatrix}0 & \sin2\theta & 2\cos2\theta & -\sin2\theta\\
-\sin2\theta & -2\cos2\theta & \sin2\theta & 0
\end{pmatrix}\\
+g\begin{pmatrix}\cos4\theta & -3\sin4\theta & -3\cos4\theta & \sin4\theta\\
-\sin4\theta & -3\cos4\theta & 3\sin4\theta & \cos4\theta
\end{pmatrix},
\end{aligned}
\label{eq:cubic-sym-functions}
\end{equation}
where the first term is radial rotationally invariant, the second
is radial, the third tangential, and the fourth is of none of these
types. Figure \ref{fig:cub-sym-basis} shows the action of each of
these terms on points on a circle and on a grid oriented according
to the symmetry axis.

\selectlanguage{english}%
Although for a given value of parameter $\theta$ the function sets
given by\foreignlanguage{american}{ \eqref{eq:quad-sym-functions}
or by \eqref{eq:cubic-sym-functions} are linear subspaces, }when
we consider the union of the sets corresponding to all the possible
values of $\theta$ we do not obtain a linear subspace. \foreignlanguage{american}{For
example, the polynomials 
\[
f_{1}(z,\bar{z})=z^{2},\,\,\,f_{2}(z,\bar{z})=iz\bar{z}
\]
are of the form \eqref{eq:quad-sym-complex} but their sum is not.
The obtainment of isotropic linear models constituted by displacement
functions with reflection symmetry is addressed in the following section.}
\end{example}

\subsection{Linear isotropic reflection-symmetric models\label{sec:Mirror-symmetric-polynomial-mode}}

The previous results can be employed to obtain a practical description
of linear isotropic quadratic models of reflection symmetric functions,
given by the following theorem, whose proof is included in the \ref{subsec:proof-srf-lin-general},
in the appendix.
\begin{theorem}
\label{th:rsf-lin-general} The linear isotropic distortion models
with monomials of degree at most $n$ constituted by functions with
reflection symmetry are those of the form 
\begin{equation}
\mathcal{M}_{m}^{(n)}\left[f,g\right]\oplus{\cal F},\label{eq:linear-isotropic-distortion-models}
\end{equation}
where the spaces $\mathcal{M}_{m}^{(n)}\left[f,g\right]$ are defined
in theorem \ref{th:irreducible-real-subspaces-all}, $f,g$ are polynomials
with real coefficients, and ${\cal F}$ is a subspace generated by
invariant monomials \eqref{eq:invariant-monomials} with real coefficients.\footnote{Note that if $f=g=0$ then $\mathcal{M}_{m}^{(n)}\left[f,g\right]=\left\{ 0\right\} $
and that $\mathcal{F}$ can also be the null vector subspace.}
\end{theorem}
\begin{example}
\label{exa:quadratic-ref-sym}As we saw in example \ref{example:irred-dim-2},
the irreducible subspaces in ${\cal P}^{(2)}$ are the spaces 
\[
\mathcal{M}_{1}^{(2)}(\mu:\nu)=\left\{ \gamma\mu z^{2}+\bar{\gamma}\bar{\nu}z\bar{z}\colon\gamma\in\mathbb{C}\right\} ,\,(\mu:\nu)\in\mathbb{P}_{\mathbb{C}}^{1}
\]
and the space 
\[
{\cal P}_{-3}^{(2)}=\mathcal{M}_{3}^{(2)}\left[\bar{z}^{2},0\right]=\left\{ \gamma\bar{z}^{2}\colon\gamma\in\mathbb{C}\right\} 
\]
and there are not invariant monomials. Therefore the linear isotropic
quadratic distortion models constituted by functions with reflection
symmetry are the spaces $\mathcal{M}_{1}^{(2)}(\mu:\nu)$ with $\mu,\nu\in\mathbb{R}$
and ${\cal P}_{3}^{(2)}$. In the first case we have, noting $\mu=r,\,\nu=s$,
$r,s\in\mathbb{R}$, and $\gamma=ae^{i\phi}$, $a,\phi\in\mathbb{R}$,
\[
\mathcal{M}_{1}^{(2)}(r:s)=\left\{ a\left(re^{i\phi}z^{2}+se^{-i\phi}z\bar{z}\right):a,\phi\in\mathbb{R}\right\} .
\]
Noting $p=r+s$, $q=s-r$, $t_{1}=a\cos\phi$, $t_{2}=a\sin\phi$,
it is easy to check that the real matrix form for these models is
\begin{equation}
p\begin{pmatrix}t_{1} & -t_{2} & 0\\
0 & t_{1} & -t_{2}
\end{pmatrix}+q\begin{pmatrix}0 & t_{2} & t_{1}\\
-t_{2} & -t_{1} & 0
\end{pmatrix},\,t_{1},t_{2}\in\mathbb{R},\label{eq:quad-lin-sym-param}
\end{equation}
 where the first term corresponds to radial distortion and the second
to tangential distortion. Therefore the different models of this family
are specified by the ratio between these two displacement terms.

The functions of the space ${\cal P}_{-3}^{(2)}$ are those of the
form
\[
f(z,\bar{z})=ae^{i\phi}\bar{z}^{2},\,\alpha,\phi\in\mathbb{R},
\]
and with the identification $t_{1}=a\cos\phi$, $t_{2}=a\sin\phi$,
have matrix form
\begin{equation}
\begin{pmatrix}t_{1} & 2t_{2} & -t_{1}\\
t_{2} & -2t_{1} & -t_{2}
\end{pmatrix},\,t_{1},t_{2}\in\mathbb{R}.\label{eq:quad-lin-sym-param-2}
\end{equation}
\end{example}
Therefore the set of linear isotropic quadratic distortion models
with functions with reflection symmetry consists in a one-parameter
family (para\-metrized by the ratio $(p:q)$) and an additional model.
All these models are two-dimensional and the ratio of their parameters,
$t_{2}/t_{1}$ determines the symmetry axis according to the relation
$t_{2}/t_{1}=\tan\phi$ for the models of the one-parameter family
and $t_{2}/t_{1}=\tan3\phi$ for the additional model.

\begin{figure}
\centering{}\includegraphics[width=1\columnwidth]{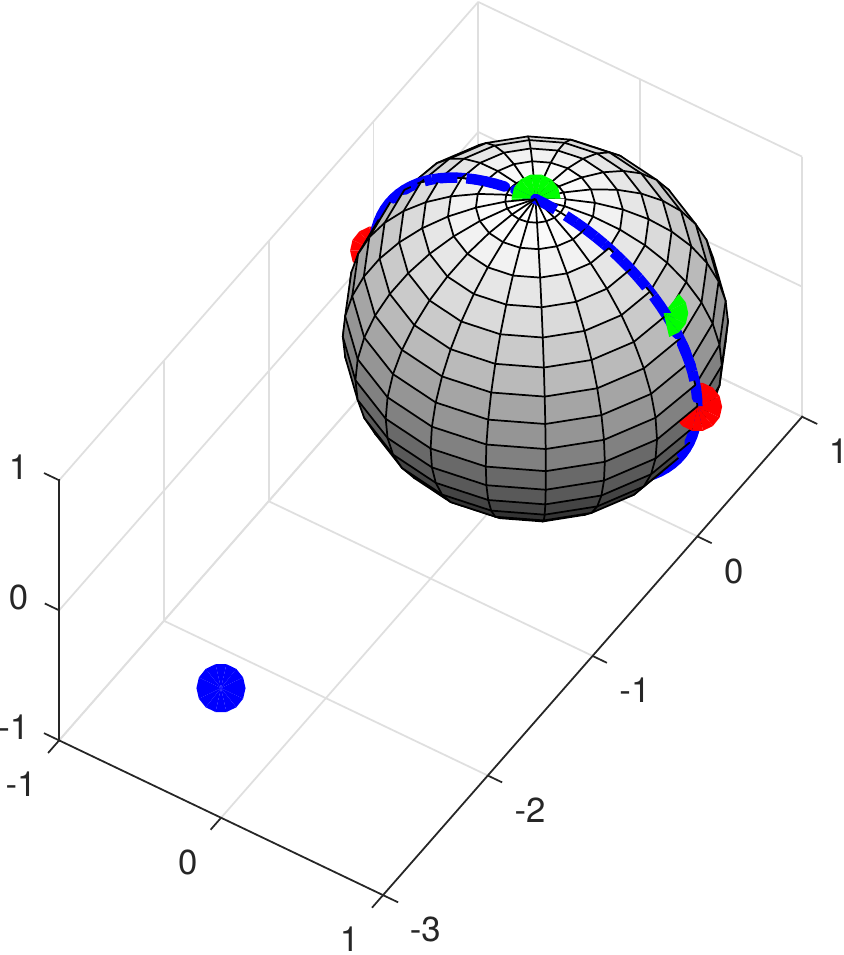}\caption{\label{fig:Topology-preserving-representation}Topology-preserving
representation of the parameter space of the irreducible isotropic
linear models of degree two (see example \ref{exa:quadratic-ref-sym}).}
\end{figure}

Figure \ref{fig:Topology-preserving-representation} provides a topology-preserving
representation of the parameter space of the irreducible isotropic
linear models of degree two. Each point of the sphere corresponds
to a bidimensional isotropic linear model $\mathcal{M}_{1}^{(2)}\left(\mu:\nu\right)$
(see equation \eqref{eq:irreducible-degree-2}) within the four-dimensional
radial-tangential space. The parameter space $\mathbb{P}_{\mathbb{C}}^{1}$
is represented as a sphere through the stereographic projection $\mathbb{P}_{\mathbb{C}}^{1}\ni(\mu:\nu)\mapsto(2\mu\bar{\nu},|\mu|^{2}-|\nu|^{2})\in\mathbb{C}\times\mathbb{R}\equiv\mathbb{R}^{3}$.
The blue circle on the sphere corresponds to those of these models
that are constituted by functions with reflection symmetry with respect
to some axis (i.e., those given by \eqref{eq:quad-lin-sym-param}),
the red dots on this circle correspond to the radial and tangential
models and the green dots correspond to the thin prism and lens decentering
models as we will see in the next section. The isolated point corresponds
to the space ${\cal P}_{-3}^{(2)}$ \eqref{eq:quad-lin-sym-param-2},
also constituted by functions with reflection symmetry. 

\section{Application: analysis of some well-known polynomial models\label{sec:Geometry-of-some-models}}

In this section we discuss how the most commonly used lens distortion
models fit in the framework presented above.

\emph{Decentering distortion} \cite{Conrady} is an analytical model
of the effect of imperfect alignment of the revolution axes of the
lens surfaces. The displacement functions of the model are given by
the quadratic functions 
\begin{equation}
\begin{aligned}\Delta x & =s_{1}\left(3x^{2}+y^{2}\right)+2s_{2}xy\\
\Delta y & =2s_{1}xy+s_{2}\left(x^{2}+3y^{2}\right).
\end{aligned}
\label{eq:decentering}
\end{equation}
In our matrix notation, the model is given by the matrices 
\[
\begin{pmatrix}3s_{1} & 2s_{2} & s_{1}\\
s_{2} & 2s_{1} & 3s_{2}
\end{pmatrix},\,\,s_{1},s_{2}\in\mathbb{R}.
\]
This model is obviously linear and, as is known from physical considerations,
it is isotropic and formed by functions with reflection symmetry.
Therefore it must be an instance of the models \eqref{eq:quad-lin-sym-param}
or \eqref{eq:quad-lin-sym-param-2}. It is easy to check that we are
in the first case, with coefficients 
\[
(p:q)=(3:1)
\]
and taking $t_{1}=s_{1}$ and $t_{2}=-s_{2}$ in \eqref{eq:quad-lin-sym-param}.

\emph{Thin prism distortion} \cite{Brown} models the effect of imperfection
in the lens manufacturing process and is given by the expression 
\begin{equation}
\begin{aligned}\Delta x & =u_{1}\left(x^{2}+y^{2}\right)\\
\Delta y & =u_{2}\left(x^{2}+y^{2}\right),
\end{aligned}
\label{eq:thin-prism}
\end{equation}
so that its matrix is 
\[
\begin{pmatrix}u_{1} & 0 & u_{1}\\
u_{2} & 0 & u_{2}
\end{pmatrix},\,\,u_{1},u_{2}\in\mathbb{R}.
\]
Observe that the displacement is always proportional to $(u_{1},u_{2})$.
We see again that this is a particular case of \eqref{eq:quad-lin-sym-param},
now corresponding to the coefficients 
\[
(p:q)=(1:1)
\]
and taking $t_{1}=s_{1}$ and $t_{2}=-s_{2}$. Therefore these two
models correspond to two points in the one-parameter family of models
defined by equation \eqref{eq:quad-lin-sym-param} as a consequence
of theorem \ref{th:rsf-lin-general}, represented as the green dots
in figure \ref{fig:Topology-preserving-representation}. 

Let us see how these models are combined in practice. The model employed
in the Matlab Computer Vision Toolbox \cite{Matlab} is the direct
sum of three-coefficient RRI distortion \eqref{eq:RRI-model} and
quadratic decentering distortion \eqref{eq:decentering} (named in
the documentation ``tangential distortion''), i.e., the model is
a particular case of \eqref{eq:linear-isotropic-distortion-models},
given by
\[
\mathcal{M}_{1}^{(2)}(1:1)\oplus{\cal G},
\]
where 
\[
\mathcal{G}=\left\{ z\left(a_{1}z\bar{z}+a_{2}z^{2}\bar{z}^{2}+a_{3}z^{4}\bar{z}^{4}\right)\colon a_{1},a_{2},a_{3}\in\mathbb{R}\right\} .
\]
Therefore, the model is composed of reflection symmetric functions.

In \cite{Weng} a four parameter model consisting in the sum of models
given by \eqref{eq:decentering} and \eqref{eq:thin-prism} is introduced.
Such a model coincides with the sum of the polynomial radial and polynomial
tangential models ${\cal P}_{1}^{(2)}\oplus{\cal P}_{-1}^{(2)}$ (see
equation \eqref{eq:quad-rad-tan}) which is then written as
\begin{equation}
{\cal P}_{1}^{(2)}\oplus{\cal P}_{-1}^{(2)}=\mathcal{M}_{1}^{(2)}(3:1)\oplus\mathcal{M}_{1}^{(2)}(1:1).\label{eq:another-decomposition}
\end{equation}

Finally we consider a more complex model employed in OpenCV 3.3 \cite{OpenCV}.
The OpenCV model substitute the polynomial RRI distortion found in
the \cite{Weng} model just considered by a rational RRI distortion
and the quadratic thin prism distortion is substituted by a quartic
expression 
\begin{equation}
\begin{aligned}\Delta x & =s_{1}r^{2}+s_{2}r^{4}\\
\Delta y & =s_{3}r^{2}+s_{4}r^{4}.
\end{aligned}
\label{eq:OpenCV-thin-prism}
\end{equation}
In order to analyze this part of the model, we observe first that
it corresponds to the complex polynomials
\begin{align*}
f(z,\bar{z}) & =\gamma_{11}z\bar{z}+\gamma_{22}z^{2}\bar{z}^{2}\\
\gamma_{11} & =s_{1}+is_{3}=\rho_{1}e^{i\theta_{1}}\\
\gamma_{22} & =s_{2}+is_{4}=\rho_{2}e^{i\theta_{2}}.
\end{align*}
Since this model has real dimension 4 and does not include invariant
monomials, it does not have the reflection symmetric property, according
to theorem \ref{th:rsf-lin-general}. To see this directly, just observe
that both monomials share the winding number $m=-1$ (see table \ref{table:monomials-by-exponent}),
but according to equations \ref{eq:rsf-general-equations}, the function
will be reflection symmetric if and only if
\[
e^{2i\theta_{1}}=e^{2i\theta_{2}},
\]
i.e., if $\theta_{1}=\pm\theta_{2}$, that requires $s_{3}/s_{1}=\pm s_{4}/s_{2}$.
Therefore the model given by \eqref{eq:OpenCV-thin-prism} does not
preserve the property of being formed of reflection symmetric functions
as one might expect for thin prism distortion.

\section{Application: extending known models\label{sec:Experiments}}

In this section we apply our results by proposing some extensions
of the usual lens distortion models and doing some preliminary testing
of them.

\begin{figure}
\begin{centering}
\includegraphics[width=1\columnwidth]{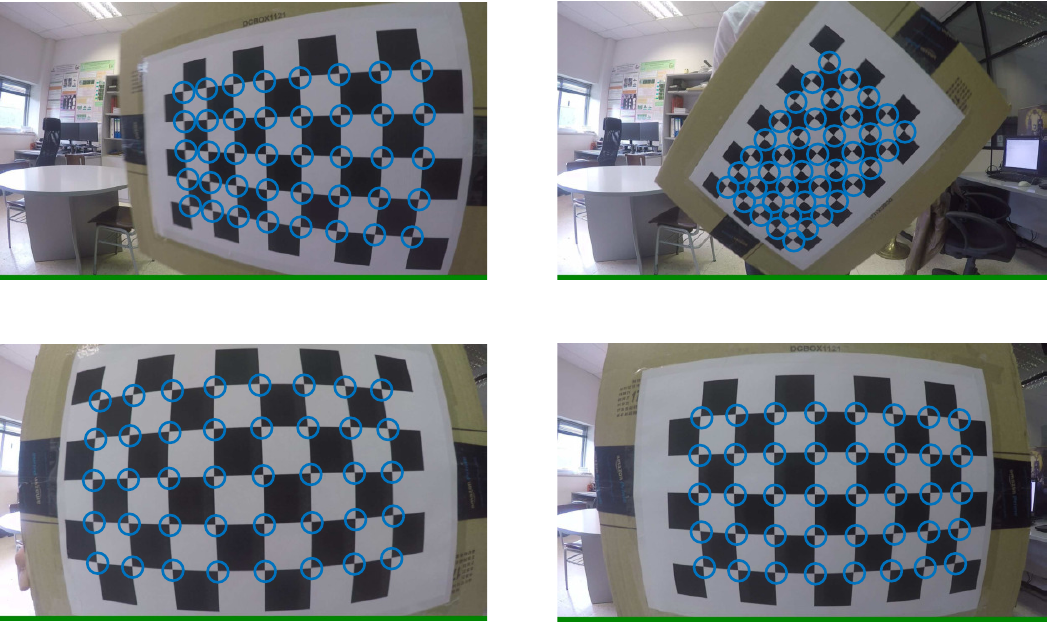} 
\par\end{centering}
\vspace{5mm}
 
\begin{centering}
\includegraphics[width=1\columnwidth]{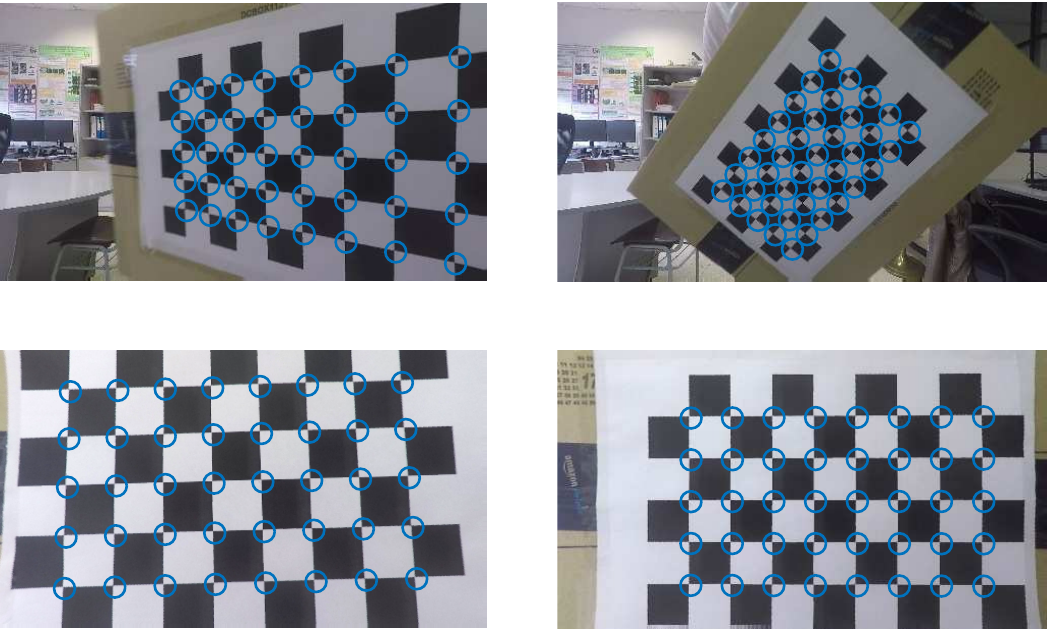} 
\par\end{centering}
\caption{\label{fig:resul}Original (four top images) and corrected (four bottom)
images with the model that minimizes reprojection error.}
\end{figure}

In order to compare different models with real images we obtain images
of a board in different positions with a GoPro camera. We first obtain
a 3D reconstruction and initial values of the distortion parameters.
For this we use the Matlab camera calibration toolbox and its model
consisting of rotationally-symmetric radial distortion of two coefficients
and quadratic decentering distortion. The distortion center is assumed
to coincide with the image principal point. Then we perform a reoptimization
of the 3D reconstruction using a different lens distortion model and
compute the residual error. Figure \ref{fig:resul} shows some original
images and their corrected versions with the best algorithm. Table
\ref{tab:Reprojection-errors} shows the reprojection errors obtained
with different models.

The first set of tests is performed with RRI distortion \eqref{eq:RRI-model}
with different number of coefficients. The improvement stops at three
coefficients. The corresponding model, which is the one generated
by the invariant monomials of degrees $3,$ 5, and 7, is kept as an
integrating part of the models considered in the remaining experiments. 

\begin{figure}
\begin{centering}
\includegraphics[width=0.9\columnwidth]{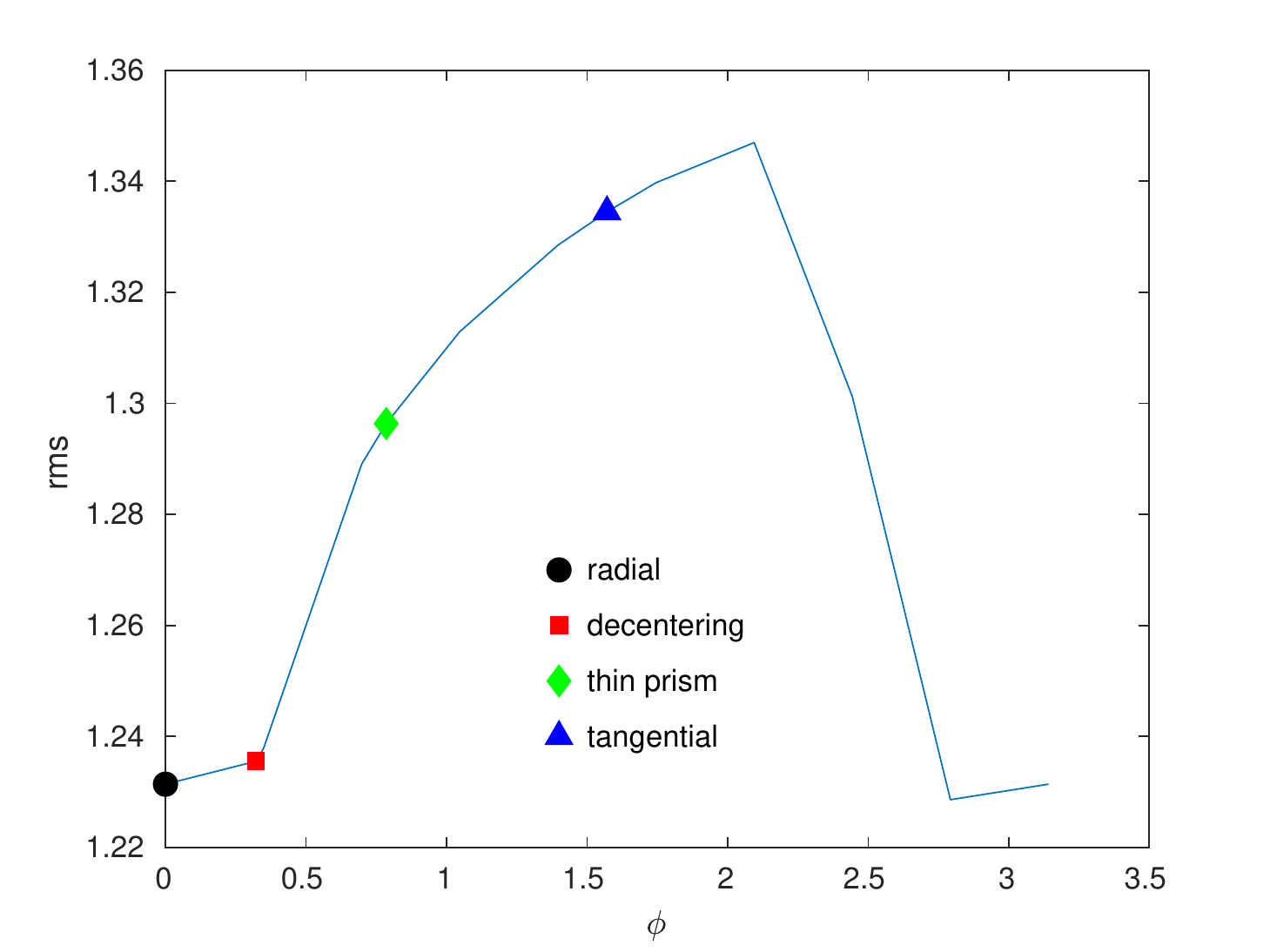} 
\par\end{centering}
\caption{\label{fig:resul-1}Residual errors (rms) for the model \eqref{eq:modelpq}
with different parameters $(p:q)=(\cos\phi:\sin\phi)$. }
\end{figure}

In the second set of experiments we consider different models of the
form 
\begin{equation}
\begin{pmatrix}\Delta x\\
\Delta y
\end{pmatrix}=\begin{pmatrix}\Delta_{1}x\\
\Delta_{1}y
\end{pmatrix}+\begin{pmatrix}\Delta_{2}x\\
\Delta_{2}y
\end{pmatrix},\label{eq:modelpq}
\end{equation}
where the first term, introduced in equation \eqref{eq:quad-lin-sym-param},
generalizes decentering and thin prism distortion and is given by
\[
\begin{pmatrix}\Delta_{1}x\\
\Delta_{1}y
\end{pmatrix}=\left[p\begin{pmatrix}t_{1} & -t_{2} & 0\\
0 & t_{1} & -t_{2}
\end{pmatrix}+q\begin{pmatrix}0 & t_{2} & t_{1}\\
-t_{2} & -t_{1} & 0
\end{pmatrix}\right]\begin{pmatrix}x^{2}\\
xy\\
y^{2}
\end{pmatrix},
\]
while the second term is the three-parameter RRI distortion \eqref{eq:RRI-model} 

\[
\begin{pmatrix}\Delta_{2}x\\
\Delta_{2}y
\end{pmatrix}=\begin{pmatrix}x\\
y
\end{pmatrix}\left(\alpha_{1}r^{2}+\alpha_{2}r^{4}+\alpha_{3}r^{6}\right).
\]
 Figure \ref{fig:resul-1} shows the residual error as a function
of the parameter $\phi$, where $(p:q)=(\cos\phi:\sin\phi)$. We observe
that the best results are achieved by models for which radial distortion
is the dominant term, i.e., for $\phi$ close to $0$ or $\pi$.

Then we consider models in which either linearity or reflection-symmetry
of the model functions is lost. First we consider linear models not
ensuring reflection-symmetry:
\begin{itemize}
\item Direct sum of decentering and thin prism distortion plus three coefficient
RRI. 
\item Full quadratic and cubic distortions with two additional coefficients
of RRI, so that the RRI term also has three coefficients in total.
\end{itemize}
Finally a nonlinear model is tested consisting in monomials of degrees
two and three ensuring reflection-symmetry (equations \eqref{eq:quad-sym-functions}
and \eqref{eq:cubic-sym-functions}), plus two additional RRI terms
in order to include three-coefficient RRI.

In table \ref{tab:Reprojection-errors} we see that the model resulting
in the minimum reprojection error is the one with largest number of
parameters, but it is closely followed by the proposed non-linear
model, that has nearly half of the parameters and enjoys the property
of being formed by reflection-symmetric functions. Therefore it seems
that for the calibration of the considered lens system the use of
models ensuring the adequate geometric properties is effective in
terms of obtaining good performance with a reduced number of parameters.

\begin{table*}
\begin{centering}
\begin{tabular}{|c|c|c|c|c|c|}
\hline 
Method  & NP  & Linear  & RRI  & RSF  & Rep. error\tabularnewline
\hline 
\hline 
1 coef. RRI & 1  & Y  & Y  & Y  & 2.71\tabularnewline
\hline 
2 coefs. RRI & 2  & Y  & Y  & Y  & 1.48\tabularnewline
\hline 
3 coefs. RRI & 3  & Y  & Y  & Y  & 1.35\tabularnewline
\hline 
4 coefs. RRI & 4  & Y  & Y  & Y  & 1.36\tabularnewline
\hline 
5 coefs. RRI & 5  & Y  & Y  & Y  & 1.36\tabularnewline
\hline 
Decentering + 3 coefs. RRI & 5  & Y  & N  & Y  & 1.24\tabularnewline
\hline 
Thin prism + 3 coefs. RRI & 5  & Y  & N  & Y  & 1.30\tabularnewline
\hline 
Radial quadratic + 3 coefs. RRI & 5 & Y  & N  & Y  & 1.23\tabularnewline
\hline 
Decentering + thin prism + 3 coefs. RRI & 7  & Y  & N  & N  & 1.20\tabularnewline
\hline 
Nonlinear quadratic and cubic + 2 extra coefs. RRI & 9  & N  & N  & Y  & 0.95\tabularnewline
\hline 
Full quadratic and cubic + 2 extra coefs. RRI  & 16  & Y  & N  & N  & 0.85\tabularnewline
\hline 
\end{tabular}
\par\end{centering}
\caption{Reprojection errors (rms) obtained after bundle adjustment with different
lens distortion models. For each model we also indicate its number
of parameters, whether it is linear, radially rotationally invariant
(RRI) and formed by reflection-symmetric functions (RSF).\label{tab:Reprojection-errors}}
\end{table*}

\section{Conclusions and future work\label{sec:Conclusions}}

In this work we have studied polynomial lens distortion models from
a geometrical point of view. After identifying the key geometrical
properties of lens distortion models, we have:
\begin{itemize}
\item provided a complete description of the models enjoying this properties,
\item placed the most commonly employed polynomial models in the resulting
picture,
\item proposed some extensions to these models enjoying the desired properties
and tested them for the calibration of a camera.
\end{itemize}
In our study we have employed the framework provided by the theory
of group representations and, to the authors knowledge, a novel representation
of polynomial models in terms of complex functions that greatly facilitates
this geometrical analysis.

Our first result has been the identification of isotropic linear models.
Then we have obtained a parameterization of the polynomial lens distortion
functions that are symmetric with respect to some axis and also the
linear isotropic models formed by functions with this property. As
an application of this result we have described all the linear quadratic
lens distortion models that are composed of reflection-symmetric functions
and found that they constitute a one-parameter family plus one particular
additional model. We have then observed that the decentering distortion
model and the thin prism model are two instances of this one parameter
family. 

Our analysis facilitates the design of polynomial models, linear or
not, enjoying the desired geometrical properties. As a practical application
of the results, some extensions of known lens distortion models have
been proposed and tested for the calibration of a camera.

A natural development of this work would be its extension to the case
of rational models.

\section{Appendix: Proofs of theorems \label{sec:Appendix}}

\subsection{Lemma to prove theorem \emph{\ref{th:irreducible-real-subspaces-all}}}
\begin{lemma}
\textbf{\label{lemma:real-invariant-subspaces}}Let $V$ be a complex
vector space with a basis $\left\{ u_{1},\ldots,u_{p},v_{1},\ldots,v_{q}\right\} $
and a complex endomorphism $f:V\to V$ given by 
\begin{align*}
f(u_{i}) & =\lambda u_{i},\,i=1,\ldots,p\\
f(v_{j}) & =\bar{\lambda}v_{j},\,j=1,\ldots,q\\
\lambda & =\lambda_{1}+i\lambda_{2}\in\mathbb{C}\setminus\mathbb{R}.
\end{align*}
Then the irreducible invariant subspaces of $f$ with respect to the
realification $V_{\mathbb{R}}$ of $V$ (i.e., the consideration of
$V$ as a real vector space by restricting the scalars to the real
numbers) are of the form
\[
{\cal S}_{(\alpha:\beta)}=\left\{ \gamma\sum_{i=1}^{p}\alpha_{i}u_{i}+\bar{\gamma}\sum_{j=1}^{q}\bar{\beta}_{j}v_{j}\colon\gamma\in\mathbb{C}\right\} ,
\]
where $(\alpha:\beta)$ is an abbreviation for 
\[
(\alpha_{1}:\ldots:\alpha_{p}:\beta_{1}:\ldots:\beta_{q})\in\mathbb{P}^{p+q-1}.
\]
Besides, if $\left(\alpha:\beta\right)\not=\left(\alpha':\beta'\right)$
then
\[
\mathcal{S}_{\left(\alpha:\beta\right)}\cap\mathcal{S}_{\left(\alpha':\beta'\right)}=\left\{ 0\right\} .
\]
\end{lemma}
\begin{svmultproof}
A basis for $V_{\mathbb{R}}$ is given by
\[
\left\{ u_{1},iu_{1},\ldots,u_{p},iu_{p},v_{1},iv_{1},\ldots,v_{q},iv_{q}\right\} ,
\]
so we can identify $V\approx\mathbb{C}^{p+q}$ and $V_{\mathbb{R}}\approx\mathbb{R}^{2(p+q)}.$
With this identification, the matrix $\mathtt{M}$ of $f$ as an endomorphism
of $\mathbb{R}^{2(p+q)}$ is block-diagonal with $p$ blocks
\[
\mathtt{B}=\begin{pmatrix}\lambda_{1} & -\lambda_{2}\\
\lambda_{2} & \lambda_{1}
\end{pmatrix}
\]
and $q$ blocks $\mathtt{B}^{\top}$. From the diagonalization
\begin{align*}
\mathbf{\mathtt{B}} & =\mathtt{U}\begin{pmatrix}\bar{\lambda} & 0\\
0 & \lambda
\end{pmatrix}\bar{\mathtt{U}}^{\top},\text{ where}\\
\mathtt{U} & =\frac{\sqrt{2}}{2}\begin{pmatrix}1 & 1\\
i & -i
\end{pmatrix},
\end{align*}
we easily obtain a diagonalization of $\mathtt{M}$ and from it we
see that the eigenvectors of this matrix associated to the eigenvalue
$\text{\ensuremath{\lambda}}$ are of the form
\begin{align}
\mathbf{w} & =\left(\alpha_{1},-i\alpha_{1},\ldots,\alpha_{p},-i\alpha_{p},\beta_{1},i\beta_{1},\ldots,\beta_{q},i\beta_{q}\right)^{\top}\nonumber \\
 & \alpha_{i},\beta_{j}\in\mathbb{C},\label{eq:M-eigenvectors}
\end{align}
and those associated to the eigenvalue $\bar{\lambda}$ are their
conjugates. Given a non null vector ${\bf w}=\mathbf{w}_{1}+i\mathbf{w}_{2}$
of this form, $\mathbf{w}$ and $\bar{\mathbf{w}}$ span an invariant
subspace of $\mathtt{M}$ whose realification admits the basis $\left\{ \mathbf{w}_{1},\mathbf{w}_{2}\right\} $.
Denoting $\alpha_{i}=a_{i}+ib_{i}$, $\beta_{j}=c_{j}+id_{j}$, we
have
\begin{align*}
\mathbf{w}_{1} & =\left(a_{1},b_{1},\ldots,a_{p},b_{p},c_{1},-d_{1},\ldots,c_{q},-d_{q}\right)^{\top}\\
\mathbf{w}_{2} & =\left(b_{1},-a_{1},\ldots,b_{p},-a_{p},d_{1},c_{1},\ldots,d_{q},c{}_{q}\right)^{\top}.
\end{align*}
The elements of this subspace have coordinates of the form
\[
r_{1}\mathbf{w}_{1}+r_{2}\mathbf{w}_{2},\,r_{1},r_{2}\in\mathbb{R},
\]
that correspond to the elements of $V$
\begin{align*}
r_{1}\left(\sum_{i=1}^{p}\underbrace{\left(a_{i}+ib_{i}\right)}_{\alpha_{i}}u_{i}+\sum_{j=1}^{q}\underbrace{\left(c_{j}-id_{j}\right)}_{\bar{\beta}_{j}}v_{j}\right)\\
+r_{2}\left(\sum_{i=1}^{p}\underbrace{\left(b_{i}-ia_{i}\right)}_{-i\alpha_{i}}u_{i}+\sum_{j=1}^{q}\underbrace{\left(d_{j}+ic_{j}\right)}_{i\bar{\beta}_{i}}v_{j}\right)\\
=\underbrace{(r_{1}-ir_{2})}_{\gamma}\sum_{i=1}^{p}\alpha_{i}u_{i}+\underbrace{(r_{1}+ir_{2})}_{\bar{\gamma}}\sum_{j=1}^{q}\bar{\beta}_{j}v_{j},
\end{align*}

and so the subspace generated by $\mathbf{w}_{1}$ and $\mathbf{w}_{2}$
is of the form ${\cal S}_{(\alpha:\beta)}$, as required. Finally,
let us see that all the irreducible subspaces are of this form. Since
$\mathtt{M}$ is real and without real eigenvectors, its irreducible
invariant subspaces are bidimensional. Therefore, let us consider
an invariant bidimensional real subspace $W\subset\mathbb{R}^{2(p+q)}\subset\mathbb{C}^{2(p+q)}$.
Let $W^{\mathbb{C}}=W\oplus iW$ be the associated complex vector
subspace. The eigenvalues of the restriction to $W^{\mathbb{C}}$
of the endomorphism given by $\mathtt{M}$ must be complex conjugated
and so they are $\left\{ \lambda,\bar{\lambda}\right\} $. The eigenvector
$\mathbf{x}=\mathbf{x}_{1}+i\mathbf{x}_{2}$ associated to the first
eigenvalue must be of the form \eqref{eq:M-eigenvectors}. The endomorphism
being real, the conjugate vector $\mathbf{\bar{x}}$ must belong to
the invariant subspace $W^{\mathbb{C}}$ and so the real vectors $\mathbf{x}_{1},\mathbf{x}_{2}\in W$
and therefore $W$ is of the form ${\cal S}_{(\alpha:\beta)}$ as
required.

As for the last assertion, just observe that if
\[
\gamma\sum_{i=1}^{m}\alpha_{i}\mathbf{u}_{i}+\bar{\gamma}\sum_{j=1}^{n}\bar{\beta}_{j}\mathbf{v}_{j}=\gamma'\sum_{i=1}^{m}\alpha'_{i}\mathbf{u}_{i}+\bar{\gamma'}\sum_{j=1}^{n}\bar{\beta}'_{j}\mathbf{v}{}_{j}
\]
then, the vectors being a base, we have that $\gamma\alpha_{i}=\gamma'\alpha'_{i}$
and $\bar{\gamma}\bar{\beta}_{j}=\bar{\gamma}'\bar{\beta}_{j}'$ and
so $(\alpha_{1}:\ldots:\alpha_{m}:\beta_{1}:\ldots:\beta_{n})=(\alpha'_{1}:\ldots:\alpha'_{m}:\beta'_{1}:\ldots:\beta'_{n})$.
\end{svmultproof}

\subsection{Proof of theorem \ref{th:rsf-lin-general} \label{subsec:proof-srf-lin-general}}

If ${\cal S}$ is a subspace of $\mathcal{P}^{(n)}$ generated by
some set of monomials and $f\in\mathcal{P}^{(n)}$, we define the
\emph{projection $P_{\mathcal{S}}(f)$ }as the polynomial obtained
by keeping in $f$ only the monomials in ${\cal S}$. Therefore, we
have a linear mapping\emph{
\[
P_{\mathcal{S}}:\mathcal{P}^{(n)}\longrightarrow\mathcal{S}.
\]
}

Now we can proceed to the proof of theorem \ref{th:rsf-lin-general}.
\begin{svmultproof}
We consider displacement functions expressed as complex polynomials
in the variables $z$ and $\bar{z}$, 
\[
f(z,\overline{z})=\sum_{(k,l)\in G^{(n)}}\gamma_{kl}z^{k}\bar{z}^{l}\in{\cal P}^{(n)}
\]
with reflection symmetry with respect to some axis. Therefore the
coefficients can be obtained through the parameterization \eqref{eq:ref-sym-param}.

Let us suppose that we have a real vector space $L$ of functions
of this form which, at the same time, is invariant under the action
of the unitary group $SO(2)$ according to \eqref{eq:cp-pol-rot-degree-2},
i.e.,
\[
\gamma_{kl}\mapsto e^{i\theta m}\gamma_{kl}.
\]

Given an element $f$ of $L$ there must exist an element $f_{0}$
of its orbit under the action of $SO(2)$ with reflection symmetry
with respect to the horizontal axis, i.e., with real coefficients
$\gamma_{kl}=a_{kl}\in\mathbb{R}$. Therefore, $L$ is determined
by its subset $L_{\mathbb{R}}$ of its elements with real coefficients. 

Denoting $m=k+l-1$ and $m'=k'+l'-1$, let us consider two pairs $(k,l)$
and $(k',l')$ such that 
\[
mm'\not=0,\,\,\text{and }|m|\not=|m'|.
\]
Let us see that $L_{\mathbb{R}}$ cannot contain a polynomial with
both coefficients $a_{kl}\neq0$ and $a_{k'l'}\neq0$. We denote by
${\cal S}$ the set of polynomials only with monomials $z^{k}\bar{z}^{l},\,z^{k'}\bar{z}^{l'}$.
Since $L$ is a linear subspace, so is its image by the linear mapping
$P_{{\cal S}}$, that cancels all monomials but $z^{k}\bar{z}^{l}$
and $z^{k'}\bar{z}^{l'}$. If such a polynomial existed, both 
\[
c\left(a_{kl}z^{k}\bar{z}^{l}+a_{k'l'}z^{k'}\bar{z}^{l'}\right)
\]
and 

\[
a_{kl}e^{i\theta m}z^{k}\bar{z}^{l}+a_{k'l'}e^{i\theta m'}z^{k'}\bar{z}^{l'}
\]
 would belong to this image for any $c,\theta\in\mathbb{R}$, so that
its sum 
\[
a_{kl}\left(c+e^{i\theta m}\right)z^{k}\bar{z}^{l}+a_{k'l'}\left(c+e^{i\theta m'}\right)z^{k'}\bar{z}^{l'}
\]
 must also be in the image, and therefore satisfy \eqref{eq:rsf-general-equations},
so that
\begin{align*}
\left(\frac{c+e^{i\theta m}}{c+e^{-i\theta m}}\right)^{2m'}=\left(\frac{c+e^{i\theta m'}}{c+e^{-i\theta m'}}\right)^{2m}
\end{align*}
for any $c,\theta\in\mathbb{R}$. If this were true we would have
that
\begin{equation}
F(z)=\left(\frac{c+z^{m}}{c+z^{-m}}\right)^{2m'}=\left(\frac{c+z^{m'}}{c+z^{-m'}}\right)^{2m}=G(z),\label{eq:analytic-equation}
\end{equation}
but 
\[
\left(\frac{d^{3}F}{dz^{3}}-\frac{d^{3}G}{dz^{3}}\right)(1)=-\frac{4\,c{\left(c-1\right)}}{{\left(c+1\right)}^{3}}{\left(m'^{2}-m^{2}\right)}mm'\not=0
\]
unless $|m|=|m'|$ or $mm'=0$, and therefore we have found a contradiction.

Let us see now that the image of $L_{\mathbb{R}}$ by the mapping
$P_{{\cal W}}$, that only keeps the non-invariant monomials of each
polynomial cannot be of dimension larger than one. It is easy to check
that a vector space is of dimension larger than one if and only if
some projection onto a coordinate plane has dimension larger than
one. In our case, this means that there are two different monomials
$z^{k}\bar{z}^{n-k},\,z^{k'}\bar{z}^{n'-k'}$ such that $L_{\mathbb{R}}$
contains polynomials 
\[
\ldots+1z^{k}\bar{z}^{l}+0z^{k'}\bar{z}^{l'}+\ldots
\]
and
\[
\ldots+0z^{k}\bar{z}^{l}+1z^{k'}\bar{z}^{l'}+\ldots
\]
with $m=k+l-1$, $m'=k+l-1$, $mm'\neq0$, and using first the isotropy
of $L$ and then its linearity, we see that $L$ must contain a polynomial
\[
\ldots+e^{i\theta m}z^{k}\bar{z}^{l}+e^{i\theta'm'}z^{k'}\bar{z}^{l'}+\ldots
\]
for any $\theta,\theta'$. And applying \eqref{eq:rsf-general-equations}
to the coefficients of these monomials we would have for all $\theta,\theta'\in\mathbb{R}$,
\begin{align*}
\left(\frac{e^{i\theta m}}{e^{-i\theta m}}\right)^{2m'} & =\left(\frac{e^{i\theta'm'}}{e^{-i\theta'm'}}\right)^{2m}\\
\Leftrightarrow e^{4i\theta mm'} & =e^{4i\theta'mm'},
\end{align*}
which is not true unless $mm'=0$.

Therefore, if $P_{{\cal W}}\left(L_{\mathbb{R}}\right)$ contains
polynomials with some monomial $z^{k}\bar{z}^{l}$ with $m=k-l-1\neq0$,
$L_{\mathbb{R}}$ must be one-dimensional and, since it can only contain
polynomials with monomials with $k-l-1\in\left\{ -m,m\right\} $,
it must be of the form 
\begin{align*}
P_{{\cal W}}\left(L_{\mathbb{R}}\right) & =\left\{ \alpha\left(f+g\right)\colon\alpha\in\mathbb{R}\right\} ,
\end{align*}
where $f\in{\cal P}_{m}^{(n)}$, $g\in{\cal P}_{-m}^{(n)}$ are polynomials
with real coefficients, so that the projection of $L${\scriptsize{}
}onto the space of non-invariant monomials is
\begin{align}
P_{{\cal W}}\left(L\right) & =\left\{ \alpha\left(e^{im\theta}f+e^{-im\theta}g\right)e^{im\theta}\colon\alpha\in\mathbb{R}\right\} ,\label{eq:Mm_space_real_coefs}
\end{align}
that corresponds to $\mathcal{M}_{m}^{(n)}\left[f,g\right]$ in \eqref{eq:Mm_uv_spaces}
with $\gamma=\alpha e^{im\theta}$.

So we have the following possibilities:

(a) If $L$ does not contain polynomials with invariant monomials,
it must of the form \eqref{eq:Mm_space_real_coefs},

(b) If $L$ only contains polynomials with invariant monomials, $L$
can be any linear subspace of invariant polynomials with real coefficients.

(c) Finally, if $L$ contains polynomials with invariant monomials
and polynomials with non-invariant monomials, since $L$ is an invariant
subspace it must contain an irreducible subspace of noninvariant monomials
that must be of the form \eqref{eq:Mm_space_real_coefs}, and only
one. Therefore $L$ must also contain its projection onto the space
of invariant polynomials, and consequently $L$ is the direct sum
of a space of the form \eqref{eq:Mm_space_real_coefs} and a linear
space of invariant polynomials with real coefficients.
\end{svmultproof}

\bibliographystyle{naturemag}
\bibliography{biblio}

\end{document}